\documentclass{article}

\usepackage{microtype}
\usepackage{graphicx}
\usepackage{booktabs} 
\usepackage{caption}
\usepackage{url}
\usepackage{amsmath,amssymb,amsfonts}
\usepackage{algorithmic}
\usepackage{algorithm}
\usepackage{letltxmacro}
\usepackage{subcaption}
\usepackage{graphicx}
\usepackage[table]{xcolor} 
\usepackage{pifont}

\usepackage{multirow}
\usepackage{caption}
\usepackage{array, booktabs}       
\usepackage{wrapfig}
\usepackage{stfloats}
\usepackage{float}

\usepackage{hyperref}

\usepackage{amsmath}
\usepackage{amssymb}
\usepackage{mathtools}
\usepackage{amsthm}
\usepackage{bbding}
\usepackage[capitalize,noabbrev]{cleveref}

\usepackage{graphicx}
\theoremstyle{plain}

\theoremstyle{definition}

\theoremstyle{remark}

\usepackage[accepted]{mlsys2025}

\mlsystitlerunning{EfQAT: An Efficient Framework for Quantization-Aware Training}

\begin{document}

\twocolumn[
\mlsystitle{EfQAT: An Efficient Framework for Quantization-Aware Training}



\mlsyssetsymbol{equal}{*}

\begin{mlsysauthorlist}
\mlsysauthor{Saleh Ashkboos}{eth,equal}
\mlsysauthor{Bram Verhoef}{axelera}
\mlsysauthor{Torsten Hoefler}{eth}
\mlsysauthor{Evangelos Eleftheriou}{axelera}
\mlsysauthor{Martino Dazzi}{axelera}
\end{mlsysauthorlist}

\mlsysaffiliation{eth}{ETH Zurich}
\mlsysaffiliation{axelera}{Axelera AI}

\mlsyscorrespondingauthor{Saleh Ashkboos}{saleh.ashkboos@inf.ethz.ch}

\mlsyskeywords{Machine Learning, MLSys}

\vskip 0.3in

\begin{abstract}
Quantization-aware training (QAT) schemes have been shown to achieve near-full precision accuracy. They accomplish this by training a quantized model for multiple epochs. This is computationally expensive, mainly because of the full precision backward pass. On the other hand, post-training quantization (PTQ) schemes do not involve training and are therefore computationally cheap, but they usually result in a significant accuracy drop. We address these challenges by proposing EfQAT, which generalizes both schemes by optimizing only a subset of the parameters of a quantized model. 
EfQAT starts by applying a PTQ scheme to a pre-trained model and only updates the most critical network parameters while freezing the rest, accelerating the backward pass.
We demonstrate the effectiveness of EfQAT on various CNNs and Transformer-based models using different GPUs. Specifically, we show that EfQAT is significantly more accurate than PTQ with little extra compute. Furthermore, EfQAT can accelerate the QAT backward pass between 1.44-1.64x while retaining most accuracy. 
\end{abstract}
]

\printAffiliationsAndNotice{\mlsysEqualContribution} 

\section{Introduction}
\label{sec:intro}
Large-scale neural networks are at the core of many advances in Artificial Intelligence. 
Model scaling is of prime importance and a key enabler in achieving ever-increasing performance on a variety of cognitive tasks. Indeed, state-of-the-art models have moved from 60M parameters in 2012~\cite{krizhevsky2017imagenet} to 540B in 2022~\cite{chowdhery2022palm}, meanwhile showing a strong correlation between the increase in capabilities of the Deep Neural Network (DNNs) models and the number of parameters. Consequently, training these increasingly accurate and general models implies increasing hardware requirements in terms of memory, compute, and energy~\cite{openai2018aiandcompute,strubell2019energy}. Given the current trend, future huge models may be infeasible to train with currently available systems.

Previous work has addressed this resource challenge by introducing methods to reduce the model size. These methods often rely on quantization~\cite{gholami2021survey} and sparsification~\cite{hoefler2021sparsity}.
The former tries to reduce the computational and memory cost of neural network training and inference by allocating a lower number of bits to different network variables (e.g. weights and activations), whereas the latter aims at retaining the most critical parameters and pruning the rest. Quantization has gained more attention as it is easier to be applied in practice with the available hardware. 
As the dynamic range of activations and weights varies considerably across the network layers, the main challenge of quantizing neural networks lies in mapping real-valued, continuous operands and parameters to a discrete integer-valued domain without significant distortion.

\definecolor{dark_green}{rgb}{0.1,0.6,0.1}
\definecolor{red_color}{rgb}{0.8,0.1,0.1}
\def\colorcross{\textcolor{red_color}{\ding{54}} }
\def\colortick{\textcolor{dark_green}{\ding{52}} }
\begin{table}
	\small
\scalebox{0.95}{
	\centering
	\begin{tabular}{lcc}
		\toprule
		Quantization Schemes & \shortstack{Near FP \\ Accuracy} & \shortstack{High \\ Performance}  \\
		\midrule
		Post-Training Quantization (PTQ)   & \colorcross   & \colortick \\
		Quantization-Aware Training (QAT) & \colortick    & \colorcross \\
		\midrule
		\textbf{EfQAT}    & \colortick   & \colortick \\
		\bottomrule
	\end{tabular}
 }
	\caption{Different quantization schemes: EfQAT achieves higher accuracy than PTQ and is faster than QAT. FP refers to the full precision model.}
	\label{tab:teaser}
\end{table}

\textbf{Post-training quantization} (\textbf{PTQ}) schemes quantize models in a greedy fashion. Specifically, these algorithms usually optimize an auxiliary loss function representing the distance between the quantized and unquantized operands and parameters in the network. The optimization is applied on a \textit{per-layer} basis using only a few training samples, known as the calibration set.
Although these approaches are computationally inexpensive, they often decrease network accuracy.

\textbf{Quantization-aware training} (\textbf{QAT}) is another class of neural network quantization algorithms that tries to make the above transformation towards lower precision networks easier by training the quantized network directly. These schemes jointly optimize the quantization- and network parameters using
standard gradient-based methods (like Stochastic Gradient Descent) on the original network loss. QAT methods have the benefit of higher accuracy compared to PTQ methods. However, QAT methods are computationally expensive as they only accelerate the forward pass, while 
the backward pass should be performed in high precision, which has twice the number of operations as the forward pass. Thus, QAT methods are not computationally efficient for large models nor are suitable for computationally-constrained environments.

To address these challenges, we generalize the above quantization schemes and introduce Efficient QAT (\textbf{EfQAT}). EfQAT starts from a quantized model (output of an arbitrary quantization scheme) and fine-tunes the quantization parameters as well as \textit{the most important} weights of the network. 
EfQAT aims to combine the benefits of QAT and PTQ as it accelerates the backward pass of the training process up to 2x compared to standard QAT schemes and it also improves the accuracy relative to PTQ schemes (Table \ref{tab:teaser}) We summarize our contributions as follows:

\begin{itemize}
	\item We propose EfQAT, a general quantization framework that optimizes the quantization-related parameters and the most critical network weights. We define a simple metric to extract such weights and estimate the importance of the weight channels (rows) in the convolution (linear) layers. Our scheme can be used on top of any PTQ and QAT scheme, improving their accuracy (for PTQ schemes) and performance (for QAT schemes).
	\item We demonstrate that, when EfQAT is applied to networks with 4-bit weights and 8-bit activations, we obtain a 3\% improvement over PTQ schemes using ResNet-50  on ImageNet, and 6 F1 on the SQuAD using  BERT$_\text{base}$) while updating only 5\% of the weights.
 
	\item We discuss the theoretical speedup of EfQAT and show that our approach can accelerate
 the backward pass of QAT schemes by up to 2x. We validate the practical performance of EfQAT on various tasks and achieve up to 1.64x speedup for convolution-based models and 1.45x for Transformer-based models compared to the quantization-aware training schemes.

\end{itemize}

\section{Related Work}
\label{sec:related}

\begin{figure*}[!t]
\vskip 0.2in
\begin{center}
\centerline{\includegraphics[width=1.8\columnwidth]{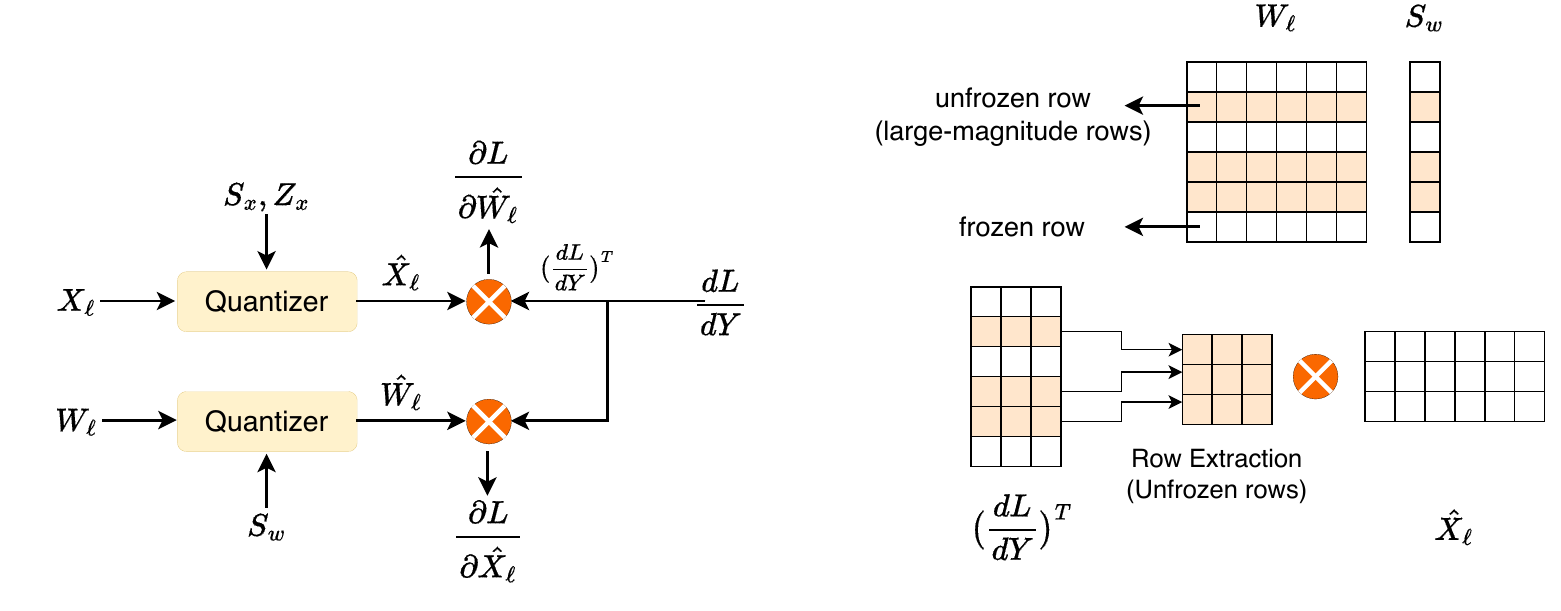}}
\caption{Main backward matrix multiplications of the quantized layer with Symmetric weight (using scale $S_w$) and Asymmetric input (using scale $S_x$ and zero point $Z_x$) quantization. 
\textbf{Left}: Quantization-aware training applies both matrix multiplications in full precision.
\textbf{Right}: EfQAT accelerates the backward pass by performing the matrix multiplication only over the most important/unfrozen rows (the rows with large average magnitude).} 
\label{fig:bwd_figure}
\end{center}
\vskip -0.2in
\end{figure*}

Quantization methods for accelerating neural network inference can be divided into two main categories~\cite{gholami2021survey}.

\paragraph{Post-Training Schemes.} Given a pre-trained model, the post-training quantization (PTQ) extract the quantization-related parameters (scales and zero points)
without (or with limited) fine-tuning of the network parameters~\cite{gptq, spqr}.
In the simplest case, according to the MinMax~\cite{krizhevsky2009learning} observer, the quantization parameters are computed based on the minimum and maximum values of weights and activations.
This approach~\cite{nagel2021white} does not need additional fine-tuning and has successfully been applied to various networks such as ResNet~\cite{he2016deep} and MobileNet~\cite{sandler2018mobilenetv2} as well as large Transformer models~\cite{vaswani2017attention}. Moreover, PTQ algorithms have been further enhanced with network-specific considerations such as keeping the outlier features in higher precision~\cite{dettmers2022llm, quik} or smoothing the activations before quantization~\cite{xiao2022smoothquant, quarot}.

More sophisticated schemes try to overcome the limitations of the post-training methods by minimizing the distance between the output of the quantized and full precision layer. For example, OMSE~\cite{choukroun2019low} uses the layer-wise MSE loss function to find the scale factors of the weights and activations, whereas OBC~\cite{frantar2022optimal} uses second-order information to optimize the above loss function. Specifically, it quantizes the weights iteratively and updates the unquantized weights at each iteration to compensate for the quantization error of the quantized weights.
Finally, GPTQ~\cite{frantar2022gptq} accelerates this process by simultaneously quantizing all the weights in each column and compressing the parameters of large models like OPT-175B~\cite{zhang2022opt} and BLOOM~\cite{laurenccon2022bigscience} into 3-4 bits.

\paragraph{Quantization-Aware Training Schemes.} Unlike post-training schemes, quantization-aware training (QAT) schemes
jointly train the model and the quantization parameters by approximating the backward gradient of the rounding operation. 
Straight Through Estimator (STE)~\cite{bengio2013estimating} is the most popular approximation for the rounding function,
 which approximates the gradient by using an identity function. 
In order to quantize CNNs,  DoReFa~\cite{zhou2016dorefa} uses the STE to approximate the gradient of
the rounding function and quantizes both weights and activations. PACT~\cite{choi2018pact} uses a trainable clipping to compress the dynamic range of the activations and reduce their quantization error. 
LSQ~\cite{esser2019learned} extends the STE approximation to make it sensitive to the distance between the full precision and quantized values. By means of this approximation, LSQ learns a scaling function for quantizing the weights and activations.
 Finally, TQT~\cite{jain2020trained} stabilizes the QAT schemes by proposing an activation function for the quantization parameters.
 SQuAT~\cite{wang2022squat} quantizes the Transformer-based models using sharpness-aware optimization~\cite{foret2020sharpness} to push the quantized weights into flat minima. This approach enables the quantization of BERT with 2 to 4 bit representation of parameters.

Our work is a generalization of both quantization schemes, in which we only optimize the most important network parameters alongside with the quantization-related parameters. 
Using our scheme, we show that with at most one epoch of EfQAT, one can obtain near-full precision accuracy on various models and datasets.

\section{EfQAT}
\label{sec:efqat}

\begin{figure*}[!t]

 \centering
     \begin{subfigure}[b]{0.45\textwidth}
         \centering
         \includegraphics[width=\textwidth]{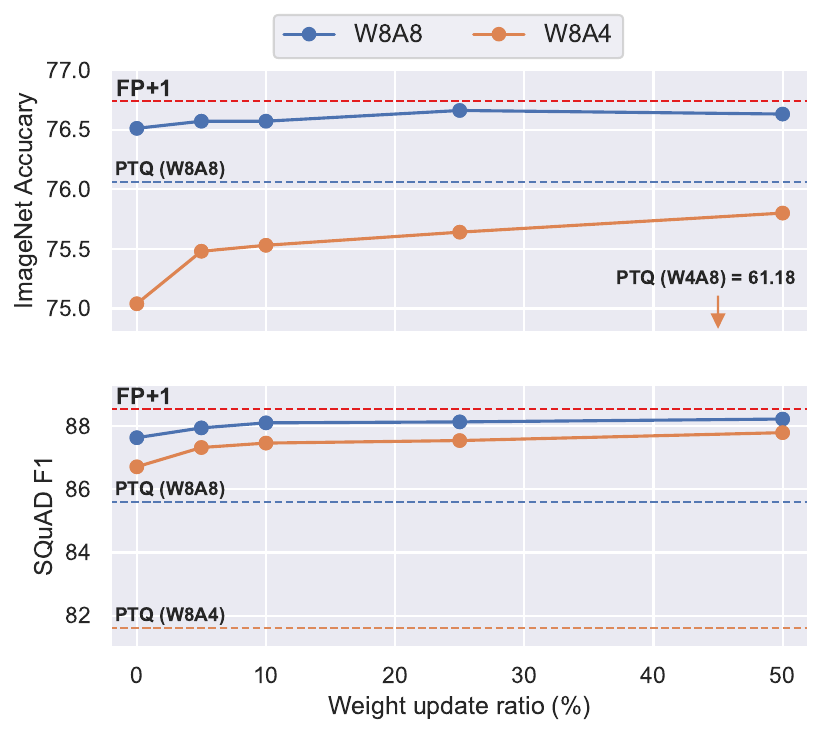}
         \caption{EfQAT-CWPN improves the accuracy over PTQ in different precisions. \textbf{FP+1} is the accuracy of training the full precision checkpoint for one 
         more epoch.}
         \label{fig:acc_highlihgt}
     \end{subfigure}
     \hfill
     \begin{subfigure}[b]{0.45\textwidth}
         \centering
         \includegraphics[width=1\textwidth]{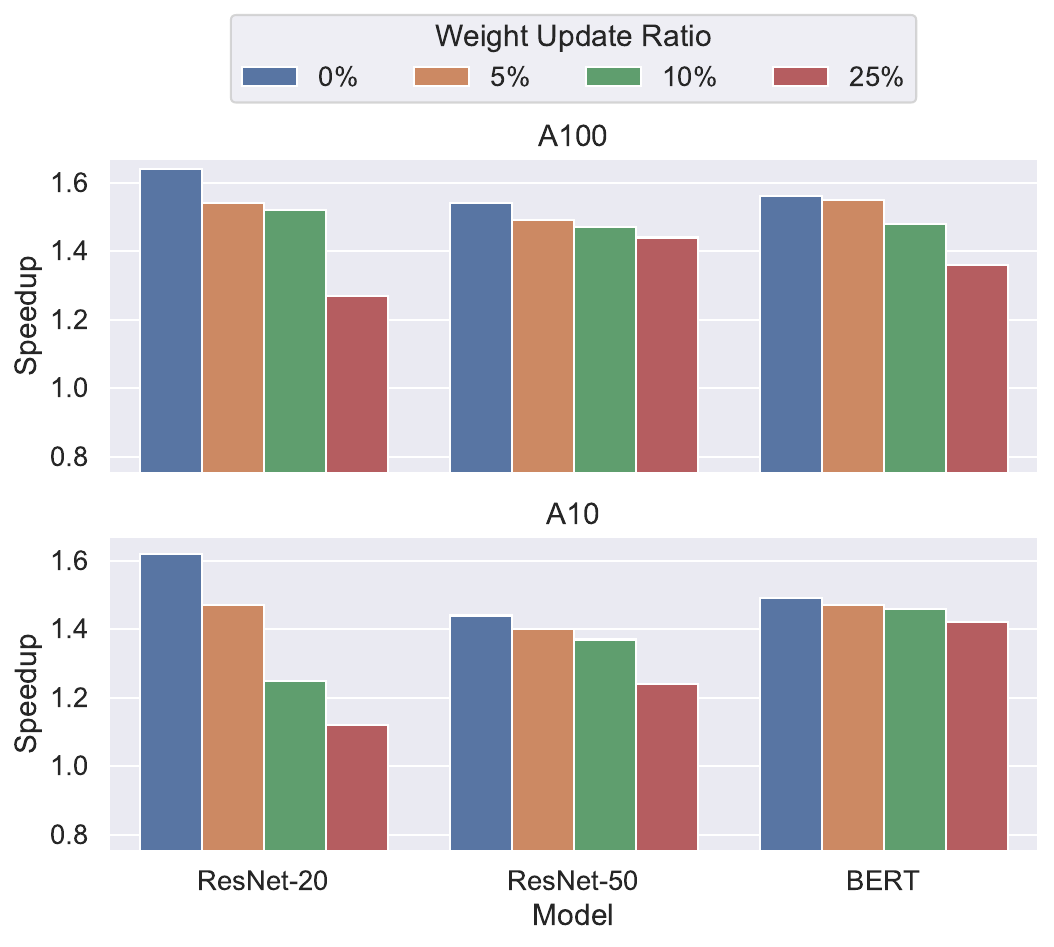}
         \caption{
         EfQAT-LWPN backward speedup over QAT on A100 and A10 GPUs. The 
         backward runtime is calculated over the total training steps during the EfQAT epoch.}
         \label{fig:speedup_highlight}
     \end{subfigure}

        \caption{Accuracy and the performance of EfQAT-CWPN/LWPN on the 
        ImageNet (using ResNet-50),  SQuAD (using  BERT$_\text{base}$), 
        and CIFAR-10 (using ResNet-20) datasets. 
        }
        \label{fig:result_highlight}

\end{figure*}

Firstly, in this section, we provide a brief introduction to quantization and discuss the main computational bottleneck of the QAT schemes.
Then, we present EfQAT and discuss its properties in detail. Finally, we provide some theoretical analysis of the speedup upper bound for our scheme.

\subsection{Background}\label{sub:Background}

\paragraph{Low-Precision Inference.}
When the distributions of the weights and activations do not have long tails (i.e., a wide dynamic range is not needed), uniformly spaced quantization points provide a practical choice~\cite{jacob2018quantization,wu2020integer}. Integer quantization provides uniform quantization that can be implemented easily in most hardware. In this work, we use the \textit{round-to-nearest} operation to transform full-precision weights and activations into low-precision values.

\paragraph{Symmetric vs. Asymmetric Quantization.}
Following \cite{nagel2021white},  we use Asymmetric Quantization for quantizing the input (activations) and Symmetric Quantization for the weights. In the former case, we round the input tensor to the $b$-bit integer using

\begin{equation}
	\hat{X} = \max \bigg(\min \bigg( \lceil \frac{X}{S_x} \rfloor + Z_x, 2^b -1 \bigg), 0 \bigg),
	\label{eq:rounding_asymm}
\end{equation}

where $X$ represents the activations, $S_x$ and $Z_x$ are the scales and zero points, known as \textit{quantization parameters}, and $\lceil . \rfloor$ is the round-to-nearest operation.  The quantization parameters are calculated using

\begin{equation}
	S_x = \frac{\beta_x - \alpha_x}{2^b -1}, \quad Z_x = - \lceil \frac{\alpha_x}{S_x} \rfloor,
	\label{eq:s_xz_x_calc}
\end{equation}

where $[ \alpha_x, \beta_x ]$ is the quantization range of the activations, estimated 
by performing the forward pass on the calibration set. We quantize the weights using

\begin{equation}
	\hat{W} = \max \bigg(\min \bigg( \lceil \frac{W}{S_w} \rfloor , -2^{b-1} + 1 \bigg), 2^{b-1} - 1 \bigg).
	\label{eq:rounding_symm}
\end{equation}

In this case, the quantization range of the weights $W$ is $[\alpha_w, \beta_w]$ and the zero points of the weights are always zero ($Z_w = 0$). The
 $S_w$ is calculated using
 
 \begin{equation}
 	S_w = \frac{ \max (|\alpha_w|, |\beta_w| )}{2^{b -1} - 1}.
 	\label{eq:s_w_calc}
 \end{equation}

\paragraph{Computational Bottleneck of QAT Schemes.}

Given a linear layer $\ell$ with weight matrix $W_{\ell}$ and input tensor $X_{\ell}$, a QAT scheme
performs the forward pass matrix-multiplication $\mathbf{Y_{\ell} = \hat{X_{\ell}}\hat{W_{\ell}}^T}$
in low precision\footnote{We follow the definition of Linear layer in the PyTorch framework~\citep{paszke2019pytorch}.}, where $\hat{X_{\ell}}$ and $\hat{W_{\ell}}$ are the quantized input and weights respectively. However, the backward pass requires two full precision
matrix multiplications (Figure \ref{fig:bwd_figure} left)

 \begin{equation}
\frac{\partial L}{\partial \hat{X_{\ell}} } = \frac{\partial L}{\partial Y_{\ell}}\hat{W_{\ell}}, \quad \quad 
\frac{\partial L}{\partial \hat{W_{\ell}}} 
 = \big(\frac{\partial L}{\partial Y_{\ell}}\big)^T\hat{X_{\ell}},
	\label{eq:qat_backward_matmuls}
\end{equation}

where $\frac{\partial L}{\partial X_{\ell}}, \frac{\partial L}{\partial Y_{\ell}}$, and $\frac{\partial L}{\partial W_{\ell}}$ are the gradients with respect to the inputs, outputs, and the weights of the layer.
These matrix multiplications dominate the computations in QAT schemes, limiting their practical applications.

\subsection{EfQAT Main Idea}\label{subsec:EfQAT_main_idea}

\paragraph{Motivation.} 
As mentioned in the previous section, the runtime of QAT schemes is dominated by two full precision matrix multiplications as in Equation (\ref{eq:qat_backward_matmuls}).
The second matrix multiplication can be accelerated by freezing a large portion of the weight matrix and only calculating the gradient for
the rest. However, this raises two significant challenges: First, we must define a metric to freeze non-important weights during the backward pass.
In addition, unstructured weight freezing does not yield practical speedup, and we need to enforce some structure on the frozen parameters. We address these challenges and define EfQAT in this section.

\paragraph{Weight Importance.}  
Inspired by previous pruning work~\cite{hoefler2021sparsity}, we define the \textit{importance} of a given block $B$ (with size $n$)  of weights from a network layer by the average magnitude of its weights. More precisely, we define 

 \begin{equation}
	\mathcal{I}_B := \frac{1}{n} \sum\limits_{w \in B} |w|,
	\label{eq:block_importance}
\end{equation}

as the metric to decide on freezing a block $B$ of weights.

\begin{figure*}[t!]
\begin{center}
\centerline{\includegraphics[width=\textwidth]{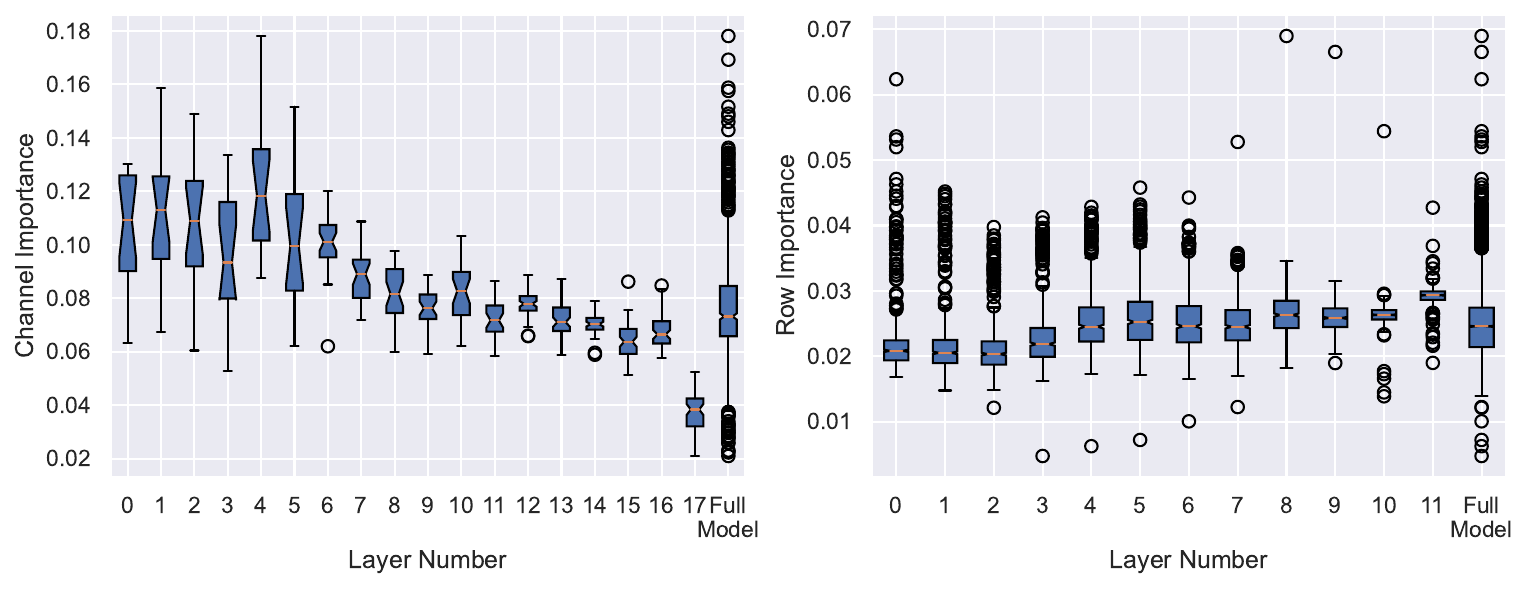}}
\caption{The importance of different channels of convolutions in ResNet-20 (left)
    and the rows of the output matrix of the self-attention layers in BERT$_\text{base}$ (right). A significant amount of outliers can be noted for both networks across different layers. In both plots, the last column displays the channel/row importance for the whole network.}
\label{fig:channel_distributions}
\end{center}
\end{figure*}

\paragraph{Structured Weight Freezing.}
Our goal is to accelerate weight gradient calculation  in Equation (\ref{eq:qat_backward_matmuls}) by only updating a small set of weights during the QAT training step.
The most natural case is to consider each channel (row) of the convolution (linear) layer as a block and freeze
the less-important weight blocks in each layer (or across the whole network). 
Figure \ref{fig:channel_distributions} shows the
$\mathcal{I}_B$ when we consider the channels (rows) in the convolution (linear) layers as the weight blocks for ResNet-20 and BERT$_\text{base}$. It suggests that there are a few important channels in different network layers.

\begin{table}[t!]
    \centering
    \vspace{0pt}
        \begin{tabular}{|c|cc|}
        \toprule
        
        \multirow{2}{*}{Mode} & \multicolumn{1}{c}{Freezing} & \multicolumn{1}{c|}{Frozen Weight} \\
        & Granularity &  Selection \\
        \midrule

        EfQAT-CWPL & Channel-Wise  & Per-Layer \\
        EfQAT-CWPN & Channel-Wise & Per-Network \\
        EfQAT-LWPN & Layer-Wise & Per-Network \\
        \bottomrule
    \end{tabular}
    \vspace{2pt}
    \caption{Three modes of EfQAT: We gradually increase the granularity of the frozen weights and the selection process.}
    \label{tab:EfQAT_modes}
\vspace{1pt}
\end{table}

\paragraph{EfQAT Modes.} Based on the granularity structure we use to freeze the network parameters, we 
define three modes for EfQAT as in Table \ref{tab:EfQAT_modes}.
In the lowest granularity level, Channel-Wise Per-Layer~(EfQAT-CWPL), we freeze the less-important \textit{channels} in each \textit{layer}. In the Channel-Wise Per-Network mode~(EfQAT-CWPN), we balance the frozen \textit{channels} across the whole \textit{network}. Finally, in the Layer-Wise Per-Network mode~(EfQAT-LWPN), we freeze the entire weights of a \textit{layer} in the \textit{network}. In all cases, we use the Top-K algorithm to find the important weights.

\paragraph{Freezing Frequency.}
The importance of the weights will change as long as we update the network parameters. However,
updating the $\mathcal{I}_B$ adds overhead to EfQAT (especially in EfQAT-CWPL and EfQAT-CWPN) as we need to iterate over all the unfrozen channels, update their importance, and then select the most important blocks. We found that we can update the frozen channels once after processing every $\mathbf{f}$ data sample without too much accuracy drop. More specifically, we see a negligible accuracy drop if we update the weight importance of every $\mathbf{f}=4096$ training sample (see Section \ref{sec:experiments}).

\paragraph{Quantization Granularity.} We use the per-channel (per-row) quantization for the weights in the convolution (linear) layers. We update the quantization parameters of only if we update the weights of that channel (unfrozen weights). We use per-tensor quantization for all the activations and apply our freezing scheme over the output channels of the convolution layer and the rows (corresponding to the output features) of the weight matrix in the linear layers.

\subsection{Algorithm Pseudocode}
Now, we present the full Pseudocode of EfQAT (EfQAT-CWPL and EfQAT-CWPN) in Algorithm \ref{alg:EfQAT_pseudocode}. The EfQAT-LWPN performs the same and we only skip the
weight gradient calculation for the frozen layers.

\begin{algorithm}[H]
	\centering
	\caption{EfQAT on a Neural Network with $L$ layers with ratio $k$.} 
	\small
	\begin{algorithmic}
		\STATE Start from a PTQ model \hfill\COMMENT{QParam initialization}
		\FOR {Each Batch $B$ in Training Set}
		\STATE Apply the Forward pass in low precision using $B$.

		\COMMENT{Backward Step:}
		\FOR {For $\ell = L,...,1$}
        \STATE Calculate the input gradient using $\frac{\partial L}{\partial \hat{X_{\ell}} } = \frac{\partial L}{\partial Y_{\ell}}\hat{W_{\ell}}$.
        \STATE Extract the unfrozen channels using $k$ and save their indices in $id$.
        \STATE Calculate the unfrozen weight gradients using 
        
        $ \frac{\partial L}{\partial \hat{W_{\ell}}} [id]  = \frac{\partial L}{\partial Y_{\ell}}^T\hat{X_{\ell}}[id]$.
		\ENDFOR
		
		\COMMENT{Optimizer Step:}
		\STATE Update the \textit{unfrozen} weight channels and their scales $S_w$.
		\STATE Update the scales and zero points of the activations $S_x, Z_x$.
		\STATE Update the biases and normalization layer.
		\ENDFOR
	
	\end{algorithmic}
	\label{alg:EfQAT_pseudocode}
\end{algorithm}

\subsection{Theoretical Speed-Up}\label{sec:theoretical_speedup}
We consider the theoretical speed-up of EfQAT-CWPL and EfQAT-CWPN with unfrozen ratio $0 \leq r \leq 1$ on two
types of layers: linear layers and convolution layers:

Consider a linear layer with the weight matrix $W \in \mathbb{R}^{C_{out}\times C_{in}}$, the input vector to the layer $X\in \mathbb{R}^{C_{in}}$, and the output gradient vector $\frac{\partial L}{\partial W_{\ell}} \in \mathbb{R}^{C_{out}} $. The total
number of operations in both backward matrix multiplications is  $2C_{in}C_{out}$.

As EfQAT calculates the gradients of $\lfloor r C_{out} \rfloor$ rows, the total number of operations can be computed as

\begin{equation}
 \begin{split}
	\text{OPS(BWD)}_{\text{Linear}} = & \underbrace{C_{in}\lfloor r C_{out} \rfloor}_{
		\text{grad. w.r.t. weights}} + C_{in}C_{out} \leq \\ &
	 (1+r) C_{in}C_{out}.
	\label{eq:ops_linear}
 \end{split}
\end{equation}

Consider a convolution layer with kernel size $k$ and matrix $\in \mathbb{R}^{C_{out}\times C_{in}\times k \times k}$. Given an input $X \in \mathbb{R}^{C_{in}\times H_{in}\times W_{in}}$, the convolution produces an output $\in \mathbb{R}^{C_{out}\times H_{out}\times W_{out}}$. Formulating the backward pass similarly to the case of the linear layer, the operation count is equal to: 

\begin{equation}
	\begin{split}
		\text{OPS(BWD)}_{\text{Conv}} = &  \underbrace{ k^{2}C_{in} \lfloor r C_{out} \rfloor H_{out}W_{out}}_{
			\text{grad. w.r.t. weights}} +  \\ &
   k^{2}C_{in}C_{out}H_{out}W_{out}   \leq  \\ & 
   (1+r) k^{2}C_{in}C_{out}H_{out}W_{out}.
		\label{eq:ops_conv}
	\end{split}
\end{equation}

The upper bounds in (\ref{eq:ops_linear}) and  (\ref{eq:ops_conv})  show that we can achieve up to 2x speedup in the backward operations in the linear and convolution layers with
$r=0$. However, in practice, the QAT backward pass has other operations (e.g., approximating the rounding operator, statistical normalization, and element-wise operations during the quantization parameter gradient calculation), which we do not count in our theoretical analysis.

The minimum and maximum speedups in this section can also be applied to the EfQAT-LWPN. In that case, we have $r \in \{ 0, 1 \}$ as we freeze all weights in a layer. However, in practice, EfQAT-CWPL and EfQAT-CWPN need to extract the rows of the matrices which has memory access (due to data movement) overhead.

\section{Experimental Results}
\label{sec:experiments}

\begin{table}[t!]
\scalebox{0.99}{
    \centering
    \vspace{0pt}
        \begin{tabular}{|c|cc|cc|}
        \toprule
        \multirow{2}{*}{Model} & \multicolumn{2}{c|}{Full Precision} & \multicolumn{2}{c|}{PTQ} \\
        & FP & FP+1 & Bit-Width & Accuracy \\
        \midrule
        
        \multirow{3}{*}{ResNet-20} &  \multirow{3}{*}{91.71} & \multirow{3}{*}{91.74} & W8A8  & 91.69 $\pm$ 0.03 \\
         &   &  & W4A8  & 88.17 $\pm$ 0.07  \\
         &   &  & W4A4  & 80.75 $\pm$ 0.41  \\
        \midrule

        \multirow{3}{*}{ResNet-50} &  \multirow{3}{*}{76.12} & \multirow{3}{*}{76.74} & W8A8  & 76.06 \\
         &   &  & W4A8  & 61.18 \\
         &   &  & W4A4  & 19.12 \\
        \midrule

        \multirow{2}{*}{BERT$_\text{base}$} &  \multirow{2}{*}{88.07} & \multirow{2}{*}{88.54} & W8A8  & 85.59  $\pm$ 0.79 \\
         &   &  & W4A8  & 81.62  $\pm$ 1.17  \\
        \bottomrule
    \end{tabular}
    }
    \vspace{1pt}
    \caption{Overview of the baseline models. We use Top-1 accuracy for the ResNet-20 (on CIFAR-10) and
    ResNet-50 (on ImageNet) and F1 score for BERT$_\text{base}$ (on SQuAD).}
    \label{tab:Baselines}
\hfill


\end{table}

\paragraph{Setup.} We evaluate EfQAT on a variety of tasks and models. For each task, we start from a full-precision pre-trained checkpoint (\textbf{FP}) and quantize it using a \textbf{PTQ} scheme. After that, we apply one epoch of \textbf{EfQAT} (CWPL, CWPN, or LWPN) to fine-tune the quantized model.
As EfQAT applies one training epoch, in addition to the PTQ model, we train the FP model for one more epoch in full precision (\textbf{FP+1}) and compare the accuracy against it.
 To update the network parameters in EfQAT, we use the same optimizer (and hyper-parameters) as FP+1 and always use Adam~\cite{kingma2014adam} to update the quantization parameters. We use a calibration dataset consisting of 512 samples in all our experiments and STE~\cite{bengio2013estimating} to approximate the gradient of the rounding function during the EfQAT training epoch.
We quantize all convolutions and linear layers (including the input, output, and shortcut layers) of the CNNs but do not quantize the embedding layer in the BERT model. For the CIFAR-10 and SQuAD datasets, we repeat all our experiments with three random seeds and present the mean and standard deviation of the outcomes. However, as we do not observe significant variation (more than 0.01) across different seeds on the ImageNet dataset, we report a single number for those experiments. We implement EfQAT using PyTorch~\cite{paszke2019pytorch} framework. We provide our code and the related instructions in Appendix~\ref{sec:appendix_code}.

\paragraph{CIFAR-10.} We evaluate EfQAT on CIFAR-10 using the ResNet-20 network from \cite{he2016deep}. We train the model for 200 epochs using SGD with momentum 0.9 and 1e-4 weight decay to extract the FP checkpoint. The initial learning rate is 0.1, and we multiply it by 0.1 at epochs 100 and 150. We use the same optimizer as FP for the network parameters (with all states and hyperparameters) during the EfQAT epoch. To update the quantization parameters, we use a learning rate 1e-6.

\paragraph{ImageNet.} We evaluate EfQAT on ImageNet using ResNet-50. We use a pre-trained model from torchvision~\cite{torchvision2016} and train the parameters using SGD with a 1e-3 learning rate. The quantization parameters are optimized with a 1e-7 learning rate.

\paragraph{Question Answering on SQuAD.} We evaluate EfQAT on BERT$_\text{base}$~\cite{devlin2018bert} for question answering task on SQuAD v1.1~\cite{rajpurkar2016squad}. To extract the FP checkpoint, we adopt the pre-trained model from the HuggingFace Transformers library~\cite{wolf2019huggingface}, and fine-tune it for two epochs using the same settings as~\cite{chen2020lottery}. We exclude 4-bit quantization of the activations and features of BERT (i.e., W4A4) as it leads to a 4.9 F1 score in the QAT experiment. Unlike QAT, we do not update the embedding layer during EfQAT and use a learning rate of 1e-6 for the quantization parameters. 

\paragraph{PTQ Baseline.} For the PTQ baseline, we use the MinMax~\cite{krizhevsky2009learning} observer for both the weights and the activations. We apply per-channel (per-row) symmetric quantization for the weights in the convolution (linear) layers and per-tensor asymmetric quantization for the activations. The summary of our baselines is presented in Table~\ref{tab:Baselines}. 

\paragraph{Hyperparameter Exploration.} We do not apply any hyperparameter optimization to our experiments. However, we evaluate EfQAT using different learning rates for the quantization parameters. In addition, following \cite{jain2020trained}, we train the logarithm of the quantization scales in place of training them directly, to evaluate the stability of the EfQAT. In all cases, we do not observe a significant accuracy drop for EfQAT. See Appendix \ref{sec:appendix_optimization_hyperparameters} for all the results and details.

\subsection{Main Results}

\paragraph{Accuracy Results.} We first study the role of EfQAT on the accuracy of various models. 
To this end, for a given weight-update ratio, we consider the three modes of EfQAT (CWPL, CWPN, and LWPN). In each case, we apply our scheme with various weight-update ratio and compare it against QAT (where we update all network parameters and quantization parameters) and PTQ.
Table \ref{tab:acc_results} summarizes our results for different networks. Interestingly, there is a jump in the accuracy in the 0\% case where we only update the quantization parameters and the computationally light-weight parameters (like biases and the normalization layers). In addition, our results show that the changes in accuracy between different EfQAT modes are negligible in almost all the cases, which shows that the acceleration provided by EfQAT-LWPN comes at a minor accuracy cost. Finally, we observe that updating 25\% of the network parameters can achieve almost the same accuracy (with $\leq$0.5\% drop) as training the full network with QAT.

\begin{table*}[t]
\centering	
	\vspace{0pt}
 \scalebox{0.8}{
 \centering
	\begin{tabular}{|c|c|c|c|ccccc|c|}

		\toprule
		\multirow{2}{*}{Model} &  \multirow{2}{*}{Bit-Width}  & \multirow{2}{*}{PTQ}  & \multirow{2}{*}{Mode}   & \multicolumn{5}{c|}{EfQAT Weight Update Ratio (\%)}  & \multirow{2}{*}{QAT}\\ 
		&  &  &  & 0 & 5 & 10 & 25 & 50  &  \\
		\midrule

        \multirow{6}{*}{BERT$_\text{base}$} & \multirow{3}{*}{W8A8} & \multirow{3}{*}{85.59  $\pm$ 0.79}  & CWPL  & \multirow{3}{*}{87.63 $\pm$ 0.04} & 87.94 $\pm$ 0.09 & 87.98 $\pm$ 0.04 & 88.05 $\pm$ 0.03 & 88.27 $\pm$ 0.02 &    \multirow{3}{*}{88.25 $\pm$ 0.12}   \\ 
         &   &    & CWPN  &  & 87.94 $\pm$ 0.07 & 88.10 $\pm$ 0.01 & 88.13 $\pm$ 0.14 & 88.22 $\pm$ 0.06 &     \\ 
         &   &    & LWPN &  &  87.81 $\pm$ 0.05 & 87.94 $\pm$ 0.19 & 88.07 $\pm$ 0.16 & 88.21 $\pm$ 0.02  &    \\ \cmidrule{2-10}

        & \multirow{3}{*}{W4A8} & \multirow{3}{*}{81.62  $\pm$ 1.17 }  & CWPL  & \multirow{3}{*}{86.71 $\pm$ 0.04} & 87.30 $\pm$ 0.04 & 87.42 $\pm$ 0.16 & 87.53 $\pm$ 0.13 & 87.82 $\pm$ 0.03  & \multirow{3}{*}{87.84 $\pm$ 0.04}  \\ 
        &   &    & CWPN  &  & 87.32 $\pm$ 0.17 & 87.46 $\pm$ 0.10 & 87.54 $\pm$ 0.06 & 87.79 $\pm$ 0.10 &    \\ 
        &   &    & LWPN &  &  87.22 $\pm$ 0.15 & 87.26 $\pm$ 0.02 & 87.56 $\pm$ 0.15 & 87.72 $\pm$ 0.12 &  \\ 
          \midrule

        \multirow{9}{*}{ResNet-20} & \multirow{3}{*}{W8A8} & \multirow{3}{*}{91.69 $\pm$ 0.03}  & CWPL  & \multirow{3}{*}{91.71 $\pm$ 0.04} & 91.72 $\pm$ 0.07 & 91.74 $\pm$ 0.04 & 91.75 $\pm$ 0.05 & 91.71 $\pm$ 0.02   & \multirow{3}{*}{91.72 $\pm$ 0.05}  \\ 
         &   &    & CWPN &  & 91.73 $\pm$ 0.01 & 91.70 $\pm$ 0.07 & 91.75 $\pm$ 0.06 & 91.72 $\pm$ 0.02 &      \\
         &   &    & LWPN  &  & 91.68 $\pm$ 0.04 & 91.69 $\pm$ 0.03 & 91.70 $\pm$ 0.01 & 91.70 $\pm$ 0.06  &   \\   \cmidrule{2-10}

        & \multirow{3}{*}{W4A8} & \multirow{3}{*}{88.17 $\pm$ 0.07}  & CWPL  & \multirow{3}{*}{91.12 $\pm$ 0.04}  & 91.14 $\pm$ 0.15 & 91.22 $\pm$ 0.07 & 91.25 $\pm$ 0.07 & 91.24 $\pm$ 0.13 & \multirow{3}{*}{91.45 $\pm$ 0.14}   \\
        &   &    & CWPN &   & 91.19 $\pm$ 0.11 & 91.16 $\pm$ 0.10 & 91.35 $\pm$ 0.11 & 91.32 $\pm$ 0.11 &      \\ 
        &   &    & LWPN &  &  91.21 $\pm$ 0.04 & 91.24 $\pm$ 0.03 & 91.32 $\pm$ 0.07 & 91.41 $\pm$ 0.10  &   \\  \cmidrule{2-10}

        & \multirow{3}{*}{W4A4} & \multirow{3}{*}{80.75 $\pm$ 0.41}  & CWPL  & \multirow{3}{*}{87.23 $\pm$ 0.34}  & 87.66 $\pm$ 0.21 & 87.59 $\pm$ 0.23 & 87.85 $\pm$ 0.31 & 87.79 $\pm$ 0.13 & \multirow{3}{*}{88.28 $\pm$ 0.37}  \\ 
        &   &    & CWPN &  & 87.50 $\pm$ 0.21 & 87.78 $\pm$ 0.19 & 87.86 $\pm$ 0.21 & 87.97 $\pm$ 0.20 &      \\ 
        &   &    & LWPN &  &  87.24 $\pm$ 0.14 & 87.60 $\pm$ 0.28 & 87.89 $\pm$ 0.07 & 88.35 $\pm$ 0.08 &    \\ 
          \midrule

        \multirow{9}{*}{ResNet-50} & \multirow{3}{*}{W8A8} & \multirow{3}{*}{76.06}  & CWPL &  & 76.65 & 76.58 & 76.65 & 76.74  &     \\ 
         &   &    & CWPN &  76.51 & 76.57 & 76.57 & 76.66 & 76.63 &  76.80  \\ 
         &   &    & LWPN &  &  76.53 & 76.51  & 76.63  & 76.68  &     \\ \cmidrule{2-10}

        & \multirow{3}{*}{W4A8} & \multirow{3}{*}{61.18}  & 
        CWPL &  & 75.31 & 75.46 & 75.72 & 75.85 &     \\ 
        &   &    & CWPN  & 75.04 & 75.48 & 75.53 & 75.64 & 75.80 & 76.02   \\ 
        &   &    & LWPN &  &  75.04 & 75.12  & 75.32  &  75.74 &     \\ \cmidrule{2-10}

        & \multirow{3}{*}{W4A4} & \multirow{3}{*}{19.12}  &  CWPL  &  & 67.41 & 67.77 & 68.32 & 68.75 &    \\ 
        &   &    &  CWPN  & 66.60 & 67.49 & 68.0 & 68.71 & 69.22 & 69.31  \\ 
        &   &    &  LWPN  &  & 67.35  & 67.7  & 68.33  & 68.14  &     \\ 
        
        \bottomrule

	\end{tabular}
	}
	\vspace{2pt}
	\caption{Accuracy of EfQAT on CIFAR-10 (using ResNet-20), ImageNet (using ResNet-50), and SQuAD v1.1 (using BERT$_\text{base}$). For each model, we evaluate different EfQAT modes, including Channel-Wise Per-Laye~(CWPL), Channel-Wise Per-Network~(CWPN), and  Layer-Wise Per-Network~(LWPN).}
	\label{tab:acc_results}
	\hfill
	
\end{table*}

\begin{figure}[t!]
\begin{center}
\centerline{\includegraphics[width=0.5\textwidth]{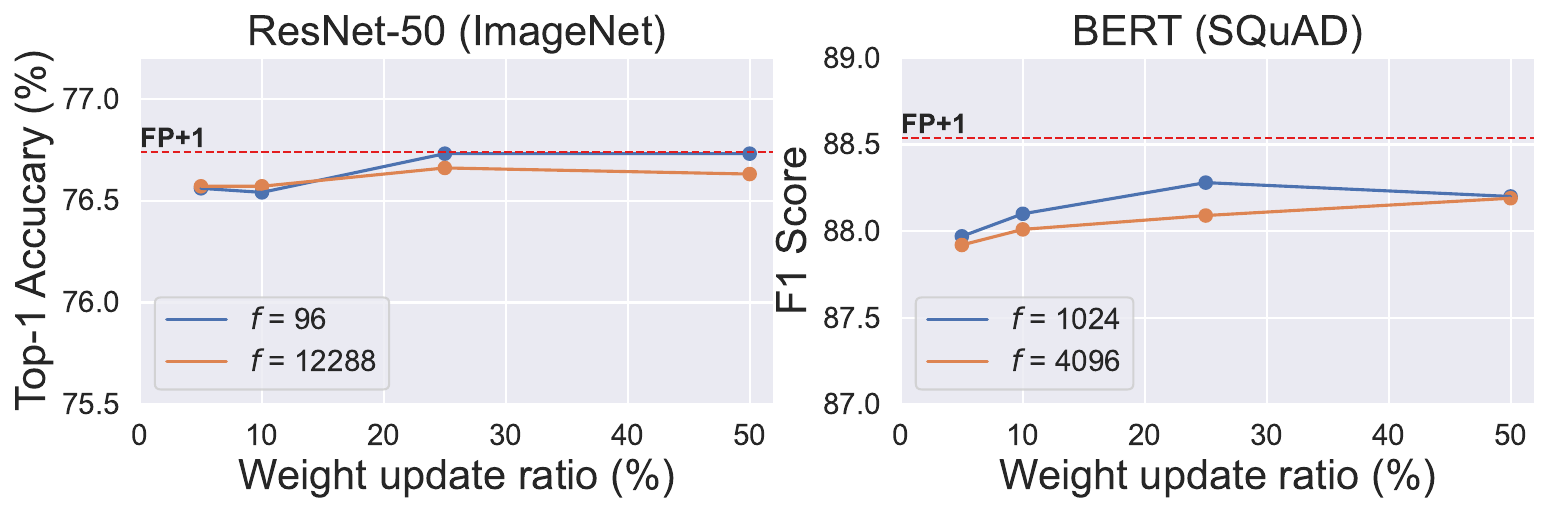}}
\caption{The role of different freezing intervals on the accuracy of EfQAT-CWPN with W8A8. We  update the frozen channels every $f$ samples. Larger $f$ does not cause a large accuracy drop during the EfQAT training epoch.}
\label{fig:freezing_freq_results}
\end{center}
\end{figure}

\paragraph{Freezing Frequency.} As mentioned in Section \ref{subsec:EfQAT_main_idea}, we need to update the frozen channels repeatedly during EfQAT training steps, which causes a computational overhead. 
To evaluate the impact of the update frequency on the accuracy, we define a frequency ratio $f$ and update the weight importances (and thus the frozen channels) for every $f>1$ training examples.
Figure \ref{fig:freezing_freq_results} shows that using large $f$ causes minor accuracy drops. Consequently, we conclude that EfQAT does not rely on frequent updates of the weight importances, making their computational overheard easily amortizable. Additional models and results are in Appendix \ref{sec:appendix_freezing_freq}.

\paragraph{Practical Speedups.} To evaluate the practical speedup of EfQAT, we run all the experiments on single A10 and A100 GPUs. We use training batch size 128 for CIFAR-10 dataset and 96 for the ImageNet dataset. For SQuAD, we use batch size 8\footnote{The largest batch size on an A10 GPU.} for the experiments on the A10 GPU and 16 for the ones on the A100. Table \ref{tab:time_bwd_per_network} compares the running time of EfQAT-CWPN and EfQAT-LWPN. We only report the speedup over the backward pass as the forward pass in QAT may be performed in low precision with hardware-dependent runtime. However, as the backward pass is the main computational bottleneck of QAT, we expect a higher end-to-end speedup when the forward pass is performed in low precision with quantized weights and activations.

EfQAT can achieve up to 1.64x speedup on A100~(1.62x speedup on A10) over the QAT by only updating the quantization parameters of the activations (0\% case). Note that we cannot obtain the full 2x theoretical speedup for two reasons: firstly, to calculate the gradients of the unfrozen weights, we need to (1) extract the corresponding rows from the output gradient matrix (Figure \ref{fig:bwd_figure} right), (2) calculate the matrix multiplication using extracted rows, and (3) finally, save the calculated gradients to the corresponding rows in the weight gradient matrix. However, the above first and last steps cause overhead due to the data movement, limiting the achievable speedup. EfQAT-LWPN solves this issue by freezing the whole weights of a layer, resulting in higher speedup. 
Secondly, the 2x theoretical speedup refers to a single convolution or linear layer. However, the backward pass of the quantized models includes other operations, such as normalizations and element-wise operations, which have been shown to constitute up to 30\% of the runtime in BERT~\cite{ivanov2021data}. While this is outside of the scope of this work, such operations could be optimized with techniques such as kernel fusion~\cite{wahib2014scalable, ben2019stateful}.

\begin{table}[t]
 \scalebox{0.73}{
	\centering
	\vspace{0pt}
	\begin{tabular}{|c|c|c|cccc|c|}

		\toprule
		Model & \multirow{2}{*}{Mode} &  Frequency & \multicolumn{4}{c|}{EfQAT Weight Update Ratio (\%)} & \multirow{2}{*}{QAT} \\ 
		(GPU)&  & ($f$) & 
		0 & 5 & 10  & 25  &  \\
		\midrule
  
		ResNet-20  & CWPN &  16384 & \multirow{2}{*}{2.61}  & 3.46   &  3.92  & 4.92  &  \multirow{2}{*}{4.23}     \\ 
        (A10) & LWPN & 16384 &   & 2.87   &  3.38  &  3.77  &    \\ 
        \midrule

         ResNet-20 & CWPN &  16384  & \multirow{2}{*}{3.73}  &  5.69  & 7.1   & 9.98 &  \multirow{2}{*}{6.22}     \\ 
        (A100) & LWPN & 16384 &   &  4.02  & 4.08   &   4.86  &    \\ 
        \midrule

		BERT$_\text{base}$  & CWPN & 4096 &  \multirow{2}{*}{1784}  & 2219   &  2271  & 2339   &   \multirow{2}{*}{2667}   \\ 
         (A10)  &  LWPN & 4096 &   & 1811   &  1821  & 1872 &     \\ 
        \midrule

		BERT$_\text{base}$  & CWPN & 4096  &  \multirow{2}{*}{1009}  &  1146   &  1173   &  1271   &   \multirow{2}{*}{1559}   \\ 
         (A100)  &  LWPN & 4096  &   &  1043   &  1055   & 1080  &     \\ 
        \midrule

        ResNet-50  & 
        CWPN & 12288 &  \multirow{2}{*}{3575}  &  6079   &  6247  & 6335 &   \multirow{2}{*}{5139}   \\ 
        (A10)   &  LWPN & 12288 &  & 3648 &  3748  & 4121 &     \\ 

        \midrule
        ResNet-50  & 
        CWPN & 12288 &  \multirow{2}{*}{1402}  &  2505    &  2545   & 2724  &   \multirow{2}{*}{2189}   \\ 
        (A100)   &  LWPN &  12288 &  &  1411 &  1474   & 1606  &     \\

        \bottomrule
	\end{tabular}
 }
	\caption{Backward runtime (in second) of EfQAT-CWPN and EfQAT-LWPN on different networks and datasets for both A100 and A10 GPUs.}
	\label{tab:time_bwd_per_network}
	\hfill
	
\end{table}

\section{Conclusion}
\label{sec:conclusion}

We propose EfQAT, a general framework to accelerate quantization-aware training schemes. EfQAT  updates only the most essential weights during the backward pass and freezes the rest. We use a simple yet efficient magnitude-based metric to choose the essential weights and freeze the rest using different granularity levels.

We apply EfQAT on different models and datasets and show that EfQAT can accelerate the QAT backward pass up to 1.64x on an A100 GPU (and 1.62x on an A10 GPU) with less than 0.3\% accuracy degradation for ResNet-50 on ImageNet and 0.62 F1 score degradation on SQuAD for BERT$_{\text{base}}$. Moreover, on the same datasets, we improve the accuracy of ResNet-50 by 0.45\%  and  of BERT$_{\text{base}}$ by 2.03 F1 score compared to PTQ.


\bibliography{References}
\bibliographystyle{mlsys2025}

\newpage
\appendix
\onecolumn
\section{SUPPLEMENTARY MATERIAL}

\subsection{Freezing Frequency}\label{sec:appendix_freezing_freq}

As mentioned in Section \ref{subsec:EfQAT_main_idea}, updating the frozen channels (rows) in the convolution (linear) layers
could be performed in large intervals. Figure \ref{fig:freezing_freq_results} shows the result of two different frequencies over the EfQAT-CWPN in the BERT$_{\text{base}}$ and ResNet-50. Table \ref{tab:freezing_freq_results} shows our results on ResNet-20 as well
on other frequencies in the BERT$_{\text{base}}$ model. We conclude that the accuracy drop is negligible across different frequencies.

\begin{table}[H]
\centering	
	\vspace{0pt}

 \centering
	\begin{tabular}{|c|c|cccc|}
 
		\toprule
		\multirow{2}{*}{Model} & Frequency & \multicolumn{4}{c|}{EfQAT Weight Update Ratio (\%)}  \\ 
	    & ($f$) & 5 & 10 & 25 & 50   \\
		\midrule

        \multirow{4}{*}{BERT$_\text{base}$} & 16 & 87.94 $\pm$ 0.07 & 88.10 $\pm$ 0.01 & 88.13 $\pm$ 0.14 & 88.22 $\pm$ 0.06 \\ 
         & 1024 & 87.97 $\pm$ 0.07 & 88.10 $\pm$ 0.16 & 88.28 $\pm$ 0.02 & 88.20 $\pm$ 0.11 \\
         & 2048 & 87.94 $\pm$ 0.04 & 88.05 $\pm$ 0.06 & 88.07 $\pm$ 0.12 & 88.21 $\pm$ 0.09 \\
         & 4096 & 87.92 $\pm$ 0.09 & 88.01 $\pm$ 0.02 & 88.09 $\pm$ 0.15 & 88.19 $\pm$ 0.14\\
         \midrule

        \multirow{3}{*}{ResNet-20} & 128 &  91.73 $\pm$ 0.01 & 91.70 $\pm$ 0.07 & 91.75 $\pm$ 0.06 & 91.72 $\pm$ 0.02 \\ 
         & 8192 & 91.73 $\pm$ 0.03 & 91.70 $\pm$ 0.04 & 91.72 $\pm$ 0.07 & 91.74 $\pm$ 0.03 \\
         & 16384 & 91.71 $\pm$ 0.05 & 91.66 $\pm$ 0.06 & 91.68 $\pm$ 0.04 & 91.72 $\pm$ 0.01 \\
         \midrule
          
        \multirow{2}{*}{ResNet-50} 
         & 96 & 76.56 & 76.54 & 76.73 & 76.73 \\
         & 12288 & 76.57 & 76.57 & 76.66 & 76.63 \\
        \bottomrule
	\end{tabular}

	\vspace{5pt}
	\caption{The role of different freezing intervals on the accuracy EfQAT-CWPN with W8A8. We  update the frozen channels every $f$ samples.}
	\hfill
	\label{tab:freezing_freq_results}
\end{table}

\subsection{Optimization Hyperparameters}\label{sec:appendix_optimization_hyperparameters}

 In the experiments discussed in Section \ref{sec:experiments}, we train the raw quantization parameters of activations and weights directly. While this is a common practice in QAT algorithms, it could lead to numerical instability. For example, while quantization scales are defined in $\mathbb{R}^{+}$, their direct optimization with a non-optimal learning rate may cause them to turn negative. In an attempt to mitigate the problem, some approaches envision the optimization of a function of the quantization parameters in place of the quantization parameters themselves. \cite{jain2020trained} proposes the training of the logarithm of the quantization scale, showing that its use together with Adam optimizer solves the numerical instability and encourages scale invariance for the updates to the quantization scales. In this section, we compare the optimization of quantization scales with and without the logarithm function. We evaluate ResNet-20 on CIFAR10 and ResNet-50 on ImageNet. In order to assess robustness to learning rate variation, we compare the performance of EfQAT with the nominal learning rate (1e-6 and 1e-7 for ResNet-20 and ResNet-50, respectively) and with a learning rate greater by a factor 1e2 for both log and raw quantization scales. Table \ref{tab:LR_study_results} reports the results. It can be noted how, for ResNet20 on CIFAR10, the training of log and raw quantization parameters appears to yield accuracy results that are all within statistical error. In the case of ResNet-50 on ImageNet, the training of raw quantization parameters with 1e-7 learning rate outperforms training with log quantization parameters.
 Although the training of the raw quantization parameters is not numerically stable, the numerical instability was not observed in our experiments even when training with significantly different learning rates. In conclusion, our experiments suggest that the performance of EfQAT is robust to variations in learning rate even when training the quantization parameters directly. Even more so, in the comparison between the two methodologies, optimization of the raw parameters always results in equivalent or higher accuracy compared to optimization of their log counterparts in out experiments.

\begin{table*}[t]
	\centering
	\vspace{0pt}
	\begin{tabular}{|c|c|c|cccc|}
		\toprule
		\multirow{2}{*}{Model} & QParam  & \multirow{2}{*}{LR} & \multicolumn{4}{c|}{EfQAT Weight Update Ratio (\%)} \\ 
		 & Func. & & 0 & 5 & 10  & 25   \\
        \midrule

         \multirow{4}{*}{ResNet-20} & \multirow{2}{*}{-}  & 1e-6 & 91.13 $\pm$ 0.08 & 91.10 $\pm$ 0.12 & 91.17 $\pm$ 0.13 & 91.32 $\pm$ 0.19  \\
          &  & 1e-4 & 91.19 $\pm$ 0.13 & 91.26 $\pm$ 0.16 & 91.34 $\pm$ 0.06 & 91.29 $\pm$ 0.09 \\\cmidrule{2-7}
          & \multirow{2}{*}{$\log_2$}  & 1e-6 &91.14 $\pm$ 0.10 & 91.11 $\pm$ 0.13 & 91.17 $\pm$ 0.15 & 91.34 $\pm$ 0.18   \\
          &  & 1e-4& 91.13 $\pm$ 0.09 & 91.26 $\pm$ 0.12 & 91.21 $\pm$ 0.11 & 91.32 $\pm$ 0.19 \\
          \midrule

          \multirow{4}{*}{ResNet-50} & \multirow{2}{*}{-}  & 1e-7 & 76.57 & 76.57 & 76.66  &  76.63 \\
          &  & 1e-5 & 75.07 & 75.41 & 75.57  &  76.20 \\\cmidrule{2-7}
          & \multirow{2}{*}{$\log_2$}  & 1e-7 & 75.08 & 75.44 &  75.38 & 75.97  \\
          &  & 1e-5& 75.03 & 75.25 &  75.44
 & 75.84  \\
        
        \bottomrule
	\end{tabular}
	\vspace{5pt}
	\caption{Accuracy of EfQAT-M2 with different learning rates and activation functions. Following \cite{jain2020trained}, we apply
    the activation functions only over the scales and train the zero points without any activation function.}
	\label{tab:LR_study_results}
	\hfill
	
\end{table*}


\end{document}